\documentclass[sigconf]{acmart}

\AtBeginDocument{%
  }

\copyrightyear{2025}
\acmYear{2025}
\setcopyright{cc}
\setcctype{by}
\acmConference[SIGIR-AP 2025]{Proceedings of the 2025 Annual International ACM SIGIR Conference on Research and Development in Information Retrieval in the Asia Pacific Region}{December 7--10, 2025}{Xi'an, China}
\acmBooktitle{Proceedings of the 2025 Annual International ACM SIGIR Conference on Research and Development in Information Retrieval in the Asia Pacific Region (SIGIR-AP 2025), December 7--10, 2025, Xi'an, China}
\acmDOI{10.1145/3767695.3769499}
\acmISBN{979-8-4007-2218-9/2025/12}

\usepackage{graphicx}
\usepackage{subcaption}
\usepackage{caption}
\usepackage[utf8]{inputenc}
\usepackage{array}
\usepackage{tabularx}
\usepackage{xcolor}
\usepackage{booktabs}
\usepackage[font=small,skip=5pt]{caption}
\usepackage{enumitem}
\usepackage{makecell}
\usepackage{multirow}
\usepackage{xcolor}  
\usepackage{balance}

\begin{document}

\title{ATACompressor: Adaptive Task-Aware Compression for Efficient Long-Context Processing in LLMs}

\author{Xuancheng Li}
\email{lixuancheng23@mails.tsinghua.edu.cn}
\authornotemark[1]
\affiliation{%
  \institution{DCST, Tsinghua University}
   \city{Beijing}
  \country{China}
}
\author{Haitao Li}
\email{liht22@mails.tsinghua.edu.cn}

\affiliation{%
  \institution{DCST, Tsinghua University}
     \city{Beijing}
  \country{China}
}

\author{Yujia Zhou}
\email{zhouyujia@mail.tsinghua.edu.cn}
\affiliation{%
  \institution{DCST, Tsinghua University}
     \city{Beijing}
  \country{China}
}

\author{Qingyao Ai}
\email{aiqingyao@gmail.com}
\affiliation{%
  \institution{DCST, Tsinghua University}
     \city{Beijing}
  \country{China}
}

\author{Yiqun Liu}
\email{yiqunliu@tsinghua.edu.cn}
\affiliation{%
  \institution{DCST, Tsinghua University}
     \city{Beijing}
  \country{China}
}

\begin{abstract}

Long-context inputs in large language models (LLMs) often suffer from the "lost in the middle" problem, where critical information becomes diluted or ignored due to excessive length. Context compression methods aim to address this by reducing input size, but existing approaches struggle with balancing information preservation and compression efficiency. We propose \textbf{A}daptive \textbf{T}ask-\textbf{A}ware Compressor (ATACompressor), which dynamically adjusts compression based on the specific requirements of the task. ATACompressor employs a selective encoder that compresses only the task-relevant portions of long contexts, ensuring that essential information is preserved while reducing unnecessary content. Its adaptive allocation controller perceives the length of relevant content and adjusts the compression rate accordingly, optimizing resource utilization. We evaluate ATACompressor on three QA datasets—HotpotQA, MSMARCO, and SQUAD—showing that it outperforms existing methods in terms of both compression efficiency and task performance. Our approach provides a scalable solution for long-context processing in LLMs. Furthermore, we perform a range of ablation studies and analysis experiments to gain deeper insights into the key components of ATACompressor\footnote{Our code is available at https://github.com/Cocobalt/ATACompressor}.
\end{abstract}



\maketitle

\section{Introduction}

Large language models (LLMs) demonstrate remarkable performance across diverse tasks, such as natural language understanding, text generation, and question answering \cite{chang2024survey,naveed2023comprehensive,min2023recent}. However, their static nature poses significant challenges. For example, they cannot independently update or adapt to new information. To bridge this gap, LLMs need external context to inject dynamic, domain-specific knowledge \cite{parthasarathy2024ultimate,wang2023survey}.
This dependency highlights the critical importance of contextual information. Without it, large models could be outdated or misaligned with real-world data, compromising both their accuracy and practical utility. 

Techniques like retrieval-augmented generation (RAG) address this challenge by retrieving relevant information from external sources, enabling the model to access up-to-date, task-specific data \cite{huang2024survey,fan2024survey}. Despite the benefits of providing ample context, naive RAG that appends raw document tokens to the model input could create excessively long context that overwhelms LLMs \cite{cuconasu2024power}, making it difficult for them to identify critical information, especially information in the middle of the context — a phenomenon commonly referred to as the "lost in the middle" problem \cite{hsieh2024found,liu2024lost}. 
\begin{figure}[!t]
    \centering
    \includegraphics[width=0.5\textwidth]{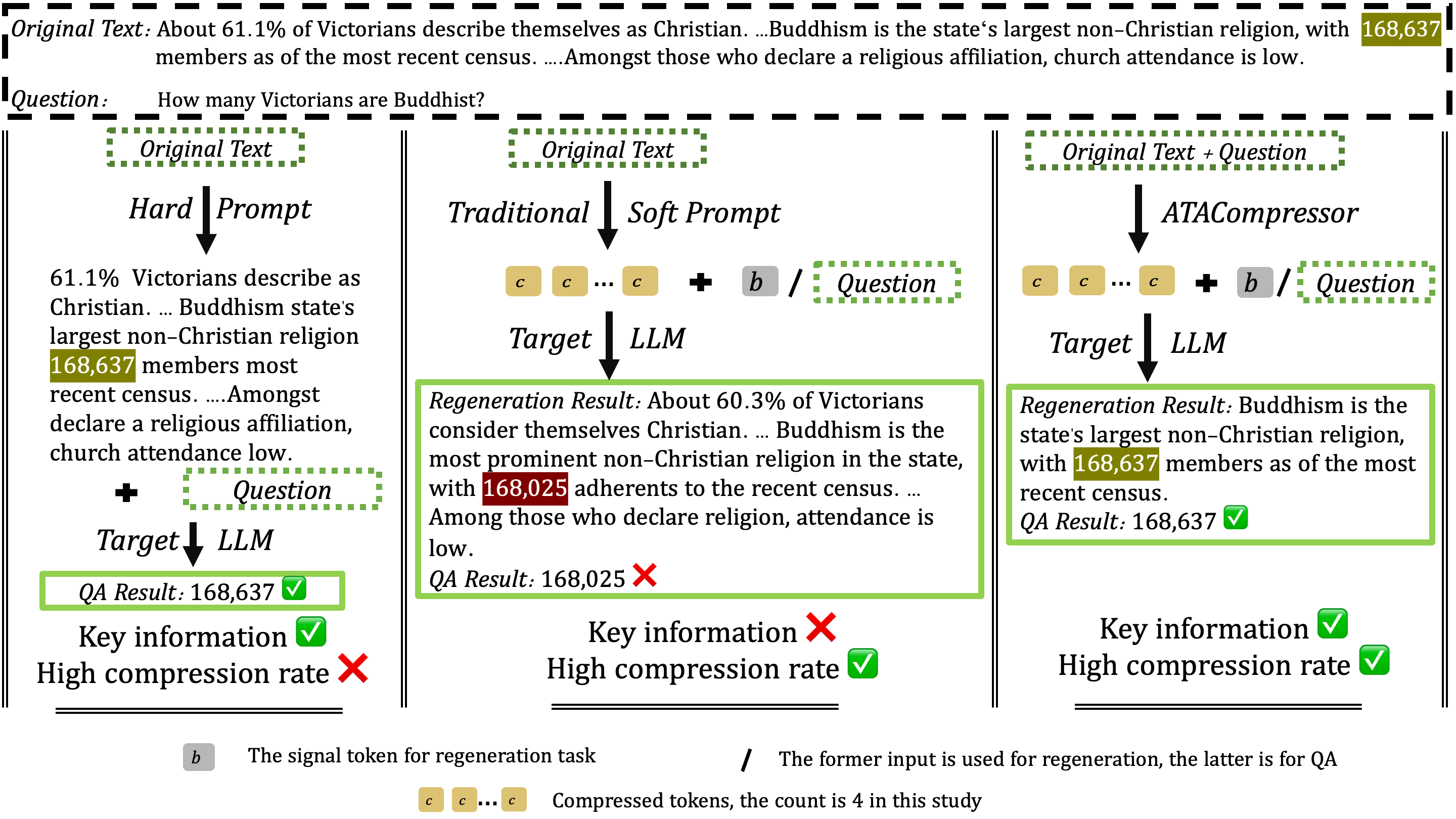} 
   \caption{Comparative schematic of three approaches, using selective compressor to represent the hard prompt and 500Compressor to represent the traditional soft prompt.}
    \label{fig:cmp}
    \vspace{-2em}
\end{figure}

One way to address this challenge is by reducing the input length. A widely adopted approach is compressing the lengthy context into a more concise form, which eases the “lost in the middle” effect and lowers inference cost and latency \cite{li2024prompt}. Existing long-context compression techniques can be broadly categorized into hard-prompt and soft-prompt methods. 
Hard prompt methods, such as Selective-Context \cite{li2023compressing} and LongLLMLingua \cite{jiang2023longllmlingua}, reduce context size by identifying and removing irrelevant or low-value content. While these methods effectively preserve task-relevant information, they typically result in lower compression ratios. 
In contrast, soft prompt methods, such as AutoCompressor \cite{chevalier2023adapting}, ICAE \cite{ge2023context}, and 500Compressor \cite{li2024500xcompressor}, compress text into a sequence of special tokens, achieving higher compression and greater information density by representing hundreds of tokens with just a few. However, these methods often suffer from the loss of task-relevant content due to the absence of task-specific information (e.g., questions) during the compression process. Additionally, their compression tokens number are fixed and cannot be dynamically adjusted based on the requirements of the task. These shortcomings prevent effective compression based on task-specific needs.

To address these challenges, we propose the \textbf{A}daptive \textbf{T}ask-\textbf{A}ware \textbf{Compressor} (ATACompressor), which offers three key advantages:
(1) \textbf{Efficient Context Compression}: ATACompressor leverages soft prompt techniques to condense long contexts into compact token representations, preserving essential information and improving downstream task efficiency.
(2) \textbf{Effective Key Information Preservation}: ATACompressor trains a selective encoder to compress only task-relevant content, filtering out irrelevant information while maximizing the retention of critical information. This task-aware compression strategy enhances downstream performance by focusing on the most important content.
(3) \textbf{Adaptive Resource Allocation}: ATACompressor employs an adaptive allocation controller that infers the length of relevant content from internal states and dynamically adjusts the compression rate accordingly. It allocates fewer tokens to shorter relevant spans and more to longer ones, ensuring adequate preservation of essential information while optimizing resource utilization across diverse tasks. Figure \ref{fig:cmp} illustrates the characteristics of ATACompressor. Our experiments on three public QA benchmarks show that ATACompressor consistently achieves state-of-the-art performance while maintaining high efficiency. 
\vspace{-1.0em}
\section{Related Work}
\subsection{Retrieval-augmented Generation}
Retrieval-Augmented Generation (RAG) enhances large language models by integrating external retrieval, improving content accuracy and factuality \cite{gao2023retrieval,zhao2024retrieval,huang2024survey,wang2023survey}. It typically combines a retrieval module with a language model to generate responses based on retrieved data \cite{liu2024lighter,gao2023retrieval,hu2024rag}.
However, RAG struggles with long contexts due to issues like the "lost in the middle" effect \cite{cuconasu2024power,liu2024lost,hsieh2024found}, where critical mid-sequence information is missed. Processing long texts also increases computational cost and latency, limiting real-time or resource-constrained use \cite{zhao2024retrieval,agrawal2024beyond}. Addressing these challenges is essential for practical long-context applications.
\vspace{-1.0em}

\subsection{Context Compression}
A common approach to handling long contexts is extending the LLM’s context window, typically via larger pretraining windows \cite{nijkamp2023xgen}, positional embedding interpolation \cite{peng2023yarn,zhu2023pose}, or attention refinements \cite{chen2023longlora}. Though effective, these methods often entail significant architectural modifications.
Unlike context extension, context compression shortens inputs without modifying LLM architecture, enabling efficient long-context handling. It consists of two types: hard and soft prompt methods. Hard methods like Selective-Context \cite{li2023compressing} and LongLLMLingua \cite{jiang2023longllmlingua} remove irrelevant tokens using external models or perplexity-based scoring but yield low compression ratios due to token retention.
Soft methods, such as AutoCompressor \cite{chevalier2023adapting}, ICAE \cite{ge2023context}, and 500Compressor \cite{li2024500xcompressor}, compress contexts into dense vectors via fine-tuning or autoencoders, achieving higher ratios but often ignoring task relevance and lacking dynamic adaptability.
Recent query-guided soft prompt methods like QGC \cite{cao2024retaining}, xRAG \cite{cheng2024xrag}, FlexRag \cite{liu2024lighter}, and COCOM \cite{rau2024context} improve task awareness but depend heavily on external retrievers and suffer from complex architectures, resulting in longer inference times and reliance on retriever quality.
Our ATACompressor algorithm, built on soft-prompt techniques, incorporates task information during compression and leverages the compressor's intrinsic ability to selectively extract, retain and compress the relevant portions of the context. It also dynamically adjusts the compression rate based on the task requirements. These features make it well-suited for RAG and other downstream tasks while delivering superior performance and efficiency.

\subsection{Probe for LLMs}
The use of probes for LLMs has gained attention for interpreting and enhancing model behavior
\cite{dong2023probing,ju2024large,ibanez2024lumia,zhao2024probe,wang2024evaluating}. 
Probing often involves attaching lightweight models to analyze and extract specific linguistic or task-related information from the LLM’s internal representations \cite{naveed2023comprehensive,wang2024exploring}.
For example, Dong et al. \cite{dong2023probing} proposes a probing mechanism to reveal both explicit and implicit gender biases in LLMs. Ju et al. \cite{ju2024large} use probing techniques to investigate the layer-wise capability of LLMs in encoding context knowledge . Wang et al. \cite{wang2024probing} leverages intrinsic probes to investigate how cross-lingual alignment emerges during the pre-training of multilingual LLMs. In our work, we apply probing techniques to analyze the hidden states of the encoder after processing the input text. This allows the probe to estimate the relevant context length, enabling a adaptive resource allocation mechanism.

\section{Method}
\begin{figure}[!t]
    \centering
    \begin{subfigure}[b]{0.5\textwidth}
        \centering
        \includegraphics[width=\textwidth]{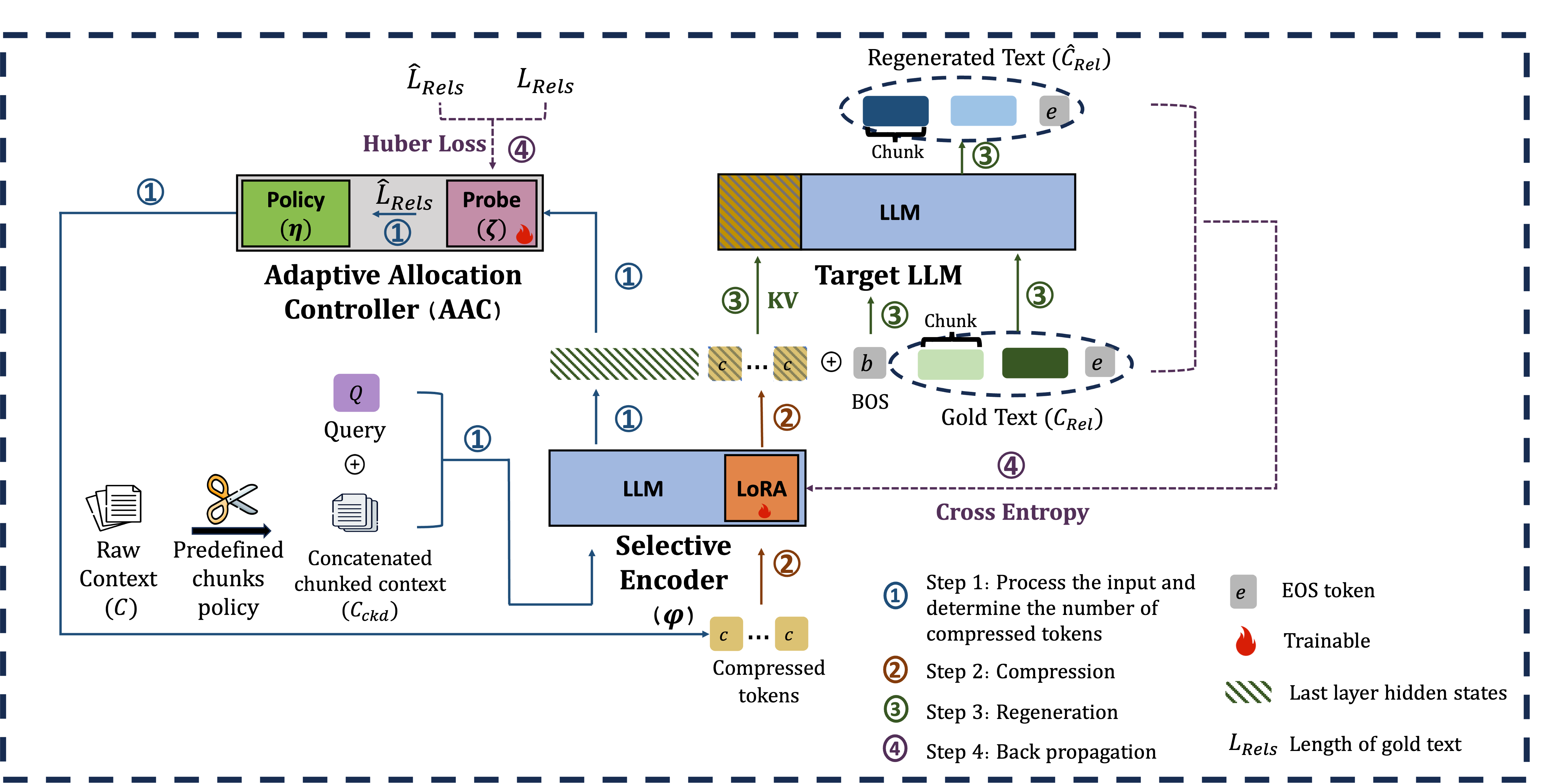}
        \caption{Pretrain}
    \end{subfigure}
    \hspace{0.05\textwidth}  
    \begin{subfigure}[b]{0.5\textwidth}
        \centering
        \includegraphics[width=\textwidth]{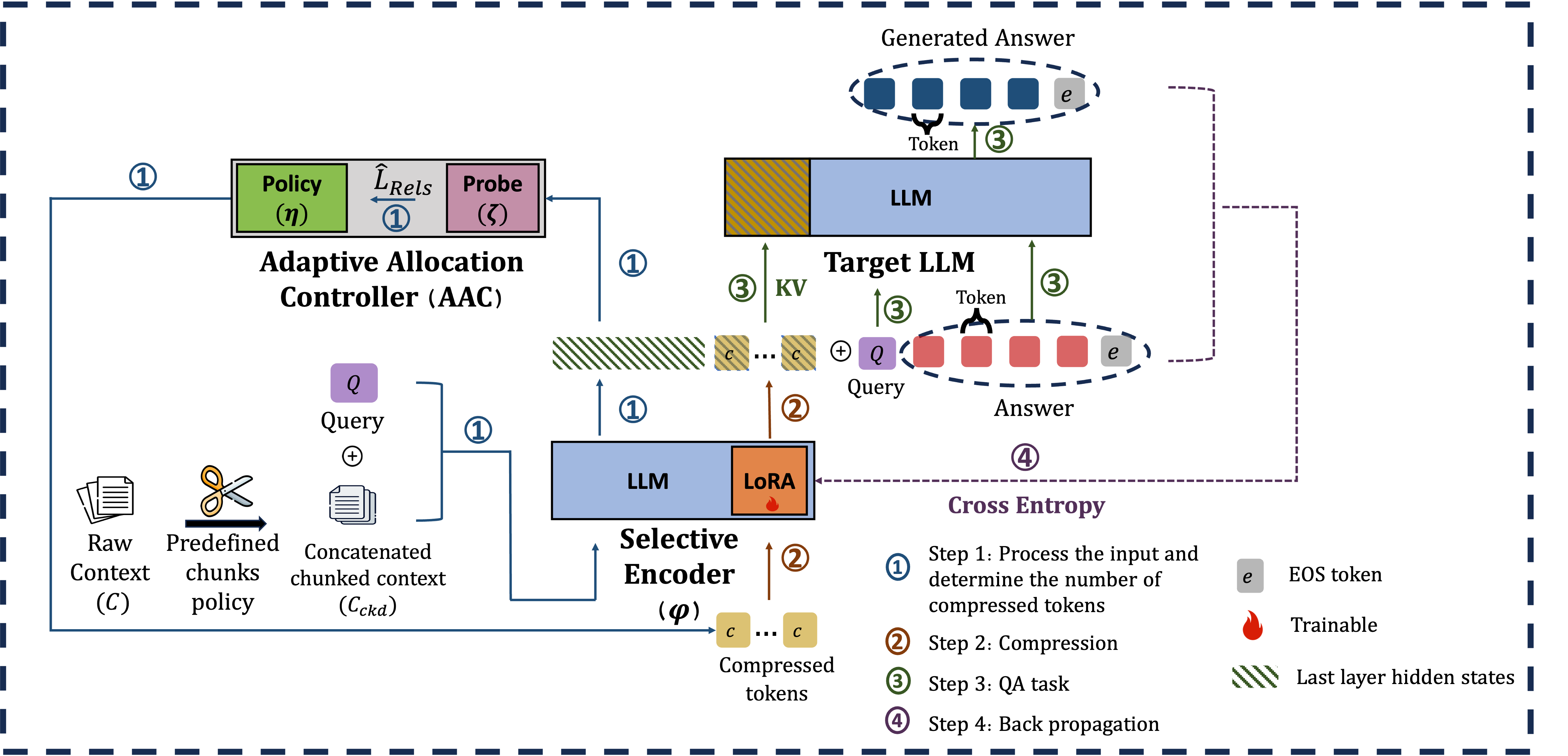}
        \caption{Finetune}
    \end{subfigure}
    
    
    \begin{subfigure}[b]{0.49\textwidth}
        \centering
        \includegraphics[width=\textwidth]{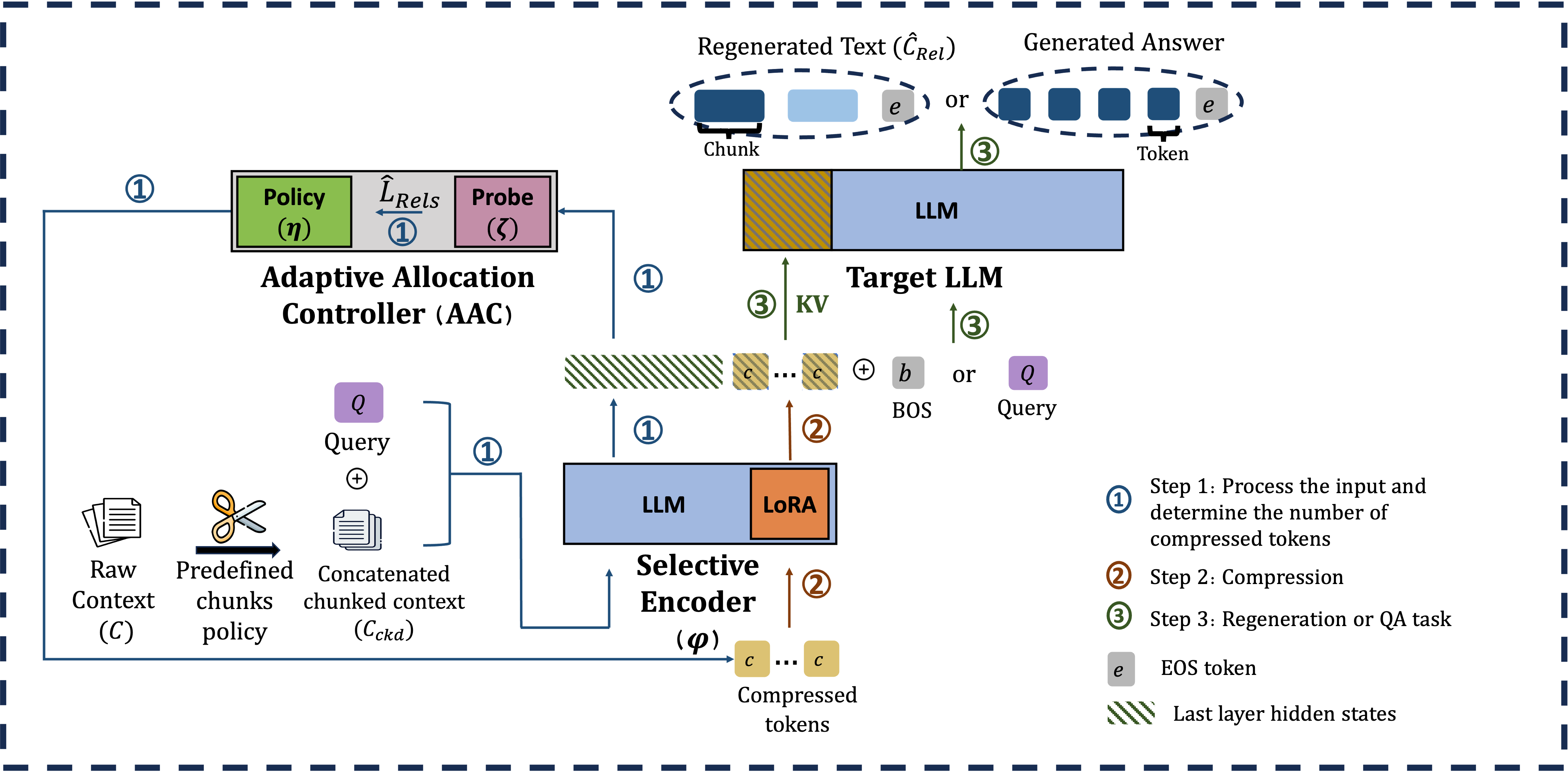}
        \caption{Inference}
    \end{subfigure}
    \caption{The architecture and workflow of ATACompressor}
    \label{fig:arch}
    \vspace{-1.5em}
\end{figure}

\subsection{Problem Formulation}
Large language models (LLMs) often take a task prompt (\(Q\)) and a context (\(C\)) as input to generate a target answer (\(A\)). However, the typically large size of \(C\) leads to challenges such as the "lost in the middle" problem, increased inference costs, longer latencies and potential performance degradation. A widely adopted way to address this challenge is context compression, the objective of ATACompressor can be formulated as:
\begin{align}
\begin{split}
\min_{\varphi(C, Q)} & \, d\big[\text{LLM}\big(A \mid C, Q\big), \text{LLM}\big(\tilde{A} \mid \varphi(C, Q), Q\big)\big] \\
\text{s.t.} \quad & |\varphi(C, Q)| = k
    \label{eq:3_1}
\end{split}
\end{align}

Here, \(\tilde{A}\) represents the output of the LLM with the compressed tokens \(\varphi(C, Q)\), \(k\) represents the number of compressed tokens, and \(d(\cdot, \cdot)\) is a distance function, such as KL divergence, that measures the difference between two distributions. 

\subsection{Architecture}
\label{3.2}
As illustrated in Figure \ref{fig:arch}, ATACompressor consists of a selective encoder, an adaptive allocation controller (AAC) and a target LLM (decoder). The general mechanism of this architecture operates as follows: first, the context \( C \) is segmented into chunks \( C_1, C_2, \dots, C_n \) using a predefined strategy. Then, the selective encoder (\( \varphi \)) processes the concatenated chunked context (denoted as \( C_{ckd} \), $C_{ckd}=\{C_1, C_2, \dots, C_n\}$) along with the query \( Q \)\footnote{In the following, we refer to all text inputs to the selective encoder as "input text".}, leveraging its inherent information sensing capability to compress the relevant portions of \(C\) required to answer \(Q\) into a set of tokens, with the number of compressed tokens \(k\) determined by the adaptive allocation controller (AAC). Subsequently, we pass the key (K) and value (V) representations of the compressed tokens to the target LLM for training or downstream tasks. It is important to note that the selective encoder (\(\varphi\)) processes the input text and compresses the relevant parts into tokens continuously, due to the autoregressive nature of LLM.
As soon as the selective encoder (\(\varphi\)) processes the last token of the input text, it continues processing and generating the first compressed token without interruption.
At this point, the AAC also begins processing in parallel. Since the AAC is very efficiently due to its relatively lightweight structure, it can predict the total number of compressed tokens before the first compressed token is generated.
Thus, although Figure \ref{fig:arch} splits the process into step 1 and step 2 for clarity,  there is no interruption between input text processing and compression in practice.

Compared to prior soft compression methods, ATACompressor introduces two key innovations: task-aware compression and dynamic token allocation. Task-aware compression is enabled by the selective encoder (\(\varphi\)), which selectively compresses only the relevant portions of the context \(C\) necessary to answer the query \(Q\), ensuring efficient use of resources while retaining crucial information. Dynamic token allocation is managed by the adaptive allocation controller (AAC), which flexibly adjusts the number of compressed tokens based on task requirements, thus preserving relevant information while optimizing computational resource usage. In the following, we will detail the design of these two components and the total workflow.

\subsection{Selective Encoder}
\label{3.3}
Traditional soft embedding methods compress the entire context \(C\) using the encoder, but they often fail to fully leverage its capacity to identify and extract relevant information. As a result, this approach may lead to the loss of content crucial for answering the query. To address this, we propose a selective encoder (\(\varphi\)) that maximizes the encoder’s ability to sense and extract relevant information. The selective encoder (\(\varphi\)) consists of a frozen large language model (LLM) \(\Theta_{\text{LLM}}\) with trainable LoRA parameters \(\Theta_{\text{LoRA}}\). It selectively retains and compresses only the portions of the context necessary to answer the query \(Q\) into a concise set of compressed tokens. This method improves compression effectiveness by preserving query-relevant content while filtering out irrelevant information, thereby enhancing the performance of downstream tasks.

A significant challenge in incorporating the selective encoder (\(\varphi\)) is that the training of traditional soft-prompt compressors \cite{ge2023context,li2024500xcompressor} is not directly applicable. In traditional pretraining, the compressor processes the entire text and aims to reconstruct it, with the gold-truth being the input itself, which is clear, consistent, and free from granularity issues. However, the selective encoder (\(\varphi\)) focuses on retaining and compressing only the relevant portions of the context required to \(Q\), meaning the gold-truth is the relevant parts of the whole context during pretraining.
This shift introduces granularity ambiguity. In real-world datasets, annotations of relevant context can vary in granularity (e.g., at the passage or document level). Without proper chunking, the inconsistent granularity of the annotated relevant context during training can confuse the selective encoder (\(\varphi\)) about how to process the context. For example, if some gold-truths indicate relevance at the document level while others are at the sentence level, the selective encoder (\(\varphi\)) will struggle with whether to group information coarsely or finely. Meanwhile, using a single granularity for training limits the selective encoder's adaptability to tasks with different granularity requirements. To address this, we segment the context into uniform chunks that align with the granularity of the corresponding gold-truth during training. The selective encoder (\(\varphi\)) is then trained to retain and compress the query-relevant chunks. This ensures the selective encoder compresses context at the corresponding granularity, even with varying granularities in the ground truths. At inference time the chunking granularity is determined by user or task requirements. Therefore, ATACompressor can effectively handle datasets with different granularities while allowing users to define more complex chunking policies for greater flexibility.

The context \(C\) is first segmented into chunks \(C_1, C_2, \dots, C_n\) using a predefined strategy. These chunks are then concatenated and processed by the selective encoder (\(\varphi\)) together with the query \(Q\). It is worth noting the following: (1) the chunking process occurs before the selective encoder's processing, meaning it is a preprocessing step rather than a task for the selective encoder (\(\varphi\)); 
(2) the chunking process is the procedure of labeling the context \( C \) according to a predefined chunking policy (e.g., using passages as the chunking unit, where each passage of \( C \) is enclosed within \texttt{<PA></PA>} tags). All the chunks are then concatenated to form a preprocessed context \( C_{ckd} \), which is input into the selective encoder (\(\varphi\)) along with the query \( Q \) and processed in a single pass rather than being processed individually in multiple passes;
(3) if the length of raw context \( C \) exceeds the selective encoder's input length limit, we can first divide \( C \) into smaller segments and then apply the selective encoder (\(\varphi\)) to compress each segment individually. This segmentation process is different from the chunking process we mentioned above.
The selective encoder (\(\varphi\)) then compresses the portions of \(C_{ckd}\) that are relevant for answering \(Q\) (denoted as \(\hat{C}_{Rel}\)) into a compact set of tokens \(c_1, c_2, \dots, c_k\), where \(k\) represents the number of compressed tokens, dynamically determined by the adaptive allocation controller (AAC). This process is formally described as follows:

{
\begin{align}
\varphi(Q,C) &=\varphi(\langle Q, C_{ckd}\rangle) =\varphi\bigl(\hat{C}_{Rel}\bigr) 
             = \underbrace{(c_1,\dots,c_k)}_{k\text{ is determined by AAC}},\\[2pt]
\hat{C}_{Rel} &= \underbrace{\{C_{t_1},\dots,C_{t_m}\}}_{ C_{t_i}\text{ is relevant to }Q}\subseteq\{C_1,\dots,C_n\},
\end{align}
\vspace{-1.5em}
}

\subsection{Adaptive Allocation Controller (AAC)}

The adaptive allocation controller (AAC) is composed of a probe (\(\zeta\)) and a policy function (\(\eta\)). The probe (\(\zeta\)) 
 captures the selective encoder's hidden states to estimate the length of relevant content. This estimation directs the policy function (\(\eta\)) to dynamically adjust the compression rate, preserving relevant information and optimizing computational resource usage.

The reason for selecting the length of \(\hat{C}_{Rel}\) as a key signal for adjusting the number of compressed tokens (\(k\)) is as follows: The length of \(\hat{C}_{Rel}\) represents length of the text needed to complete the task. Prior studies \cite{ge2023context,li2024500xcompressor,cao2024retaining,rau2024context} have demonstrated that the performance of a soft-prompt compressor is primarily influenced by the ratio between the text length and the number of compressed tokens (\(k\)) allocated during compression. Under a fixed number of compressed tokens, the performance of the compressor declines rapidly as the text length increases. By accurately estimating the length of \(\hat{C}_{Rel}\), it becomes possible to dynamically adjust the number of compressed tokens based on the requirement of the task: fewer tokens are allocated for shorter relevant texts, while more tokens are allocated for longer relevant texts.
This task-driven dynamic adjustment method maintains a balanced ratio, enabling efficient compression and enhanced performance.

The probe (\(\zeta\)) is a lightweight neural network composed of a MLP and multiple attention layers. Its primary function is to capture and analyze the hidden states produced by the selective encoder (\(\varphi\)) after processing the input text. By detecting and analyzing these hidden states, the probe (\(\zeta\))  estimates the length of \(\hat{C}_{Rel}\), which serves as the basis for dynamically adjusting the compression rate. The workflow of probe (\(\zeta\)) is as follows: 
After the selective encoder (\(\varphi\)) processes the last token of the input text, the probe (\(\zeta\)) takes the last layer hidden states of the selective encoder (\(\varphi\)) as input and outputs an estimated length \(\hat{L}_{Rel}\).  \(\hat{L}_{Rel}\) is then passed to the policy function (\(\eta\)), which dynamically determines the number of compressed tokens (\(k\)). The process can be formalized as:
\begin{equation}
    \hat{L}_{Rel} = \zeta(\mathbf{H}_{\varphi}), k= \eta(\hat{L}_{Rel})
\end{equation}

where \(\zeta(\cdot)\) represents the probe's operation on the encoder's last layer hidden states \(\mathbf{H}_{\varphi}\) to estimate \(\hat{L}_{Rel}\). 
The policy function  (\(\eta\)) determines the number of compressed tokens \(k\) based on the length of \(\hat{C}_{Rel}\) estimated by the probe (\(\zeta\)). 
In our experiments, we adopted a intuitive but effective policy: the number of compressed tokens \(k\) is determined by dividing the effective text length \(\hat{L}_{Rel}\) by a policy ratio \(r\). Specifically, \(k\) is computed as:  
\begin{equation}
k = \eta(\hat{L}_{Rel}) = \min\bigg(\frac{\hat{L}_{Rel}}{r}, k_{max}\bigg)
\label{eq:policy}
\end{equation}

where \(\hat{L}_{Rel}\) represents the length of \(\hat{C}_{Rel}\) estimated by the probe (\(\zeta\)), \(r\) is a predefined policy ratio, and \(k_{max}\) is the maximum allowable number of compressed tokens. 

It is important to note that the selective encoder ($\varphi$) and allocation controller (AAC) operate independently, ensuring that the probe ($\zeta$) structure  or policy function ($\eta$) does not affect the encoder’s ability to extract and compress information. This independence allows flexibility in designing the adaptive allocation controller, particularly the policy function ($\eta$), to suit task-specific needs. For instance, the policy ratio $r$ in our function can be adjusted without retraining, unlike traditional soft prompt methods \cite{ge2023context,li2024500xcompressor}, which require retraining to modify the number of compressed tokens.

\subsection{Workflow}

As shown in Figure \ref{fig:arch}, the workflow for ATACompressor consists of two main process: training (includes pretrain and finetune) and inference.

\subsubsection{Pretrain}
During pretraining, we jointly optimize the selective encoder (\(\varphi\)) and the probe (\(\zeta\)). The selective encoder (\(\varphi\)) is trained to extract and effectively compress the relevant portions of the context \(C\) required to query \(Q\), while the probe  (\(\zeta\)) is trained to accurately predict the relevant text length \(\hat{L}_{Rel}\).
The overall loss function for this stage combines the encoder loss (\(\mathcal{L}_{\varphi}\)) and the probe loss (\(\mathcal{L}_{\zeta}\)), with a weighting factor \(\lambda\) to balance their contributions. The total loss is defined as:
\begin{equation}
\mathcal{L}_{\text{pretrain}} = \mathcal{L}_{\varphi} + \lambda \cdot \mathcal{L}_{\zeta}
\label{eq:pretrain}
\end{equation}

\textbf{Encoder Loss (\(\mathcal{L}_{\varphi}\))}\quad
The encoder loss is formulated using a cross-entropy objective to measure the alignment between the predicted token distribution and the gold-truth token distribution. 
$C_{Rel}$ is a context formed by concatenating chunks from the context $C$ that are explicitly labeled as relevant for addressing the task $Q$, representing the annotated gold truth of $\hat{C}_{Rel}$. We ensure that the chunk granularity of $\hat{C}_{Rel}$ is consistent with that of $C_{Rel}$.
During training, teacher forcing \cite{ge2023context,li2024500xcompressor} is used to guide the LLM in reconstructing the gold-truth sequence by providing true tokens as input, enhancing the model's ability to predict the correct sequence. The loss is defined as follows:
\begin{equation}
\mathcal{L}_\varphi = -\sum_{j=1}^n \log P(w_j \mid \mathbf{KV}, [\text{BOS}], w_{1:j-1}; \Theta_{\text{LLM}}, \Theta_{\text{LoRA}})
\end{equation}
Where $w_j$ is the $j$-th token in the \(C_{Rel}\), and $\mathbf{KV}$ is the key-value representations of the compressed tokens generated by the selective encoder ($\varphi$), passed to the target LLM. The sequence starts with the beginning-of-sequence token $[\text{BOS}]$ as a signal for pretraining, while $\Theta_{\text{LLM}}$ denotes the frozen parameters of the target LLM, and $\Theta_{\text{LoRA}}$ the trainable parameters of the LoRA adapter in the selective encoder. $P(w_j \mid \cdot)$ represents the predicted probability distribution of the $j$-th token. The sequence ends when the model generates the end-of-sequence token $[\text{EOS}]$, implicitly included in the token generation process.

\textbf{Probe Loss (\(\mathcal{L}_{\zeta}\))}\quad The probe loss is calculated using the huber loss \cite{gokcesu2021generalized}, which measures the error between the estimated length \(\hat{L}_{Rel}\) and the gold-truth length \(L_{Rel}\) (length of \(C_{Rel}\)), $\delta$ is a hyperparameter:
\begin{equation}
\mathcal{L}_{\zeta}  = 
\begin{cases} 
\frac{1}{2} (\hat{L}_{Rel} - L_{Rel})^2 & \text{if } |\hat{L}_{Rel} - L_{Rel}| \leq \delta, \\
\delta (|\hat{L}_{Rel} - L_{Rel}| - \frac{\delta}{2}) & \text{otherwise}
\end{cases}
\label{eq:huber}
\end{equation}

\subsubsection{Finetune}

During the finetuning, the selective encoder (\(\varphi\)) is further trained for downstream tasks. The target LLM generates task-specific outputs based on the key-value representations of the compressed tokens. This process also employs teacher forcing. The loss function during finetuning is defined as:
\begin{equation}
    \mathcal{L}_F = -\sum_{j=1}^{n}\log P(a_j \mid \mathbf{KV}, q_{1:m}, a_{1:j-1}; \Theta_{\text{LLM}}, \Theta_{\text{LoRA}})
\end{equation}

where \(a_j\) denotes the $j$-th gold-truth answer token for the task, and \(q_{1:m}\) is the query \(Q\).

\subsubsection{Inference}

During inference, all components' parameters are frozen. As illustrated in \S\ref{3.2}, the concatenated chunked context along with the query are compressed by the selective encoder into  a set of compressed tokens. The number of compressed tokens is determined by the adaptive allocation controller. And then, the key and value representations of the compressed tokens are passed to the target LLM to generate outputs in two situations: the regeneration of \(C_{Rel}\) (triggered by the [BOS] token) and the generation of answers based on the query.
For regeneration, the target LLM predicts each token \(\hat{w}_i\) in the sequence using the probability distribution conditioned on the compressed representations and previously generated tokens:
\begin{equation}
\hat{w}_i = \arg\max_{\hat{w}_i} P(\hat{w}_i \mid \mathbf{KV}, [\text{BOS}], \hat{w}_{1:i-1}; \Theta_{\text{LLM}})
\end{equation}

For generating answers, the target LLM produces each token \(\hat{a}_j\) in the task-specific output conditioned on the compressed representations, the input query \(q_{1:m}\), and previously generated answer tokens:
\begin{equation}
\hat{a}_j = \arg\max_{\hat{a}_j} P(\hat{a}_j \mid \mathbf{KV}, q_{1:m}, \hat{a}_{1:j-1}; \Theta_{\text{LLM}})
\end{equation}

\begin{table*}[ht]
\centering
\small
\caption{The experimental results on three benchmark datasets include the following metrics: EM (Exact Match), F1 (F1 score), CR (Compression Ratio), and TP (Throughput in \textit{examples / second}). Our ATACompressor algorithm shows significant improvements in all metrics across all datasets compared to QGC and 500Compressor , with p < 0.001.}
\begin{tabular}{@{}lcccccccccccc@{}}
\toprule
\textbf{Methods} & \multicolumn{4}{c}{\textbf{HotpotQA}} & \multicolumn{4}{c}{\textbf{MSMARCO}} & \multicolumn{4}{c}{\textbf{SQUAD}} \\
\cmidrule(lr){2-5} \cmidrule(lr){6-9} \cmidrule(lr){10-13}
 & F1 & EM & CR & TP & F1 & EM & CR & TP & F1 & EM & CR & TP \\
\midrule
\midrule
{\textbf{Qwen-2-7B}} \\
\midrule
Closed-book & 30.58 & 10.30 & -- & \textbf{5.64} & 15.52 & 1.01 & -- & \textbf{2.83} & 38.85 & 6.30 & -- & \textbf{5.29} \\
Original-Context & 59.88 & 39.73 & 1.0x & 1.24 & 40.79 & 4.25 & 1.0x & 0.41 & 68.52 & 48.75 & 1.0x & 2.04\\
\midrule
Selective-Context & 53.70 & 37.08 & 3.73x & 1.24 & 32.38 & 2.55 & 4.12x & 0.66 & 59.70 & 40.50 & 4.94x & 1.64\\
LongLLMLingua & 64.60 & 40.05 & 4.67x & 1.28 & 43.55 & 4.95 & 5.92x & 0.75 & 64.85 & 48.15 & 5.36x & 2.10\\
\midrule
ICAE  & 65.35 & 39.68 & 23.16x & 3.60 & 46.23 & 5.15 & 14.58x & 1.25 & 61.30 & 45.30 & 21.48x & 2.95 \\
500Compressor & 67.40 & 42.15 & 23.16x & 3.51 & 47.20 & 5.30 & 14.58x & 1.24 & 64.65 & 47.98 & 21.48x & 2.94\\
QGC  & 72.36 & 51.50 & 13.98x & 1.83 & 49.80 & 6.10 & 16.72x & 0.79 & 66.80 & 49.25 & 17.23x & 1.49 \\


\midrule
ATACompressor & \textbf{80.23} & \textbf{65.49} & \textbf{23.81x} & 3.63 & \textbf{53.30} & \textbf{8.15} & \textbf{25.32x} & 1.35 & \textbf{70.52} &\textbf{52.10} &\textbf{27.39x} & 3.07 \\
\midrule
\midrule
{\textbf{LLaMA-2-7B}} \\
\midrule
Closed-book & 22.80 & 4.80 & -- & \textbf{6.37} & 10.96 & 0.72 & -- & \textbf{3.47} & 37.90 & 5.40 & -- & \textbf{5.67} \\
Original-Context & 53.71 & 36.20 & 1.0x & 1.21 & 38.72 & 4.09 & 1.0x & 0.44 & 68.89 & 50.38 & 1.0x & 2.12\\
\midrule
Selective-Context  & 51.65 & 36.05 & 3.52x & 1.21 & 30.60 & 2.34 & 3.83x & 0.61 & 57.69 & 39.35 & 4.78x & 1.35\\
LongLLMLingua & 62.80 & 37.30 & 4.22x & 1.31 & 43.01 & 3.47 & 4.82x & 0.86 & 65.38 & 48.20 & 4.98x & 1.71 \\
\midrule
AutoCompressor & 59.64 & 32.00 & 12.23x & 2.95 & 33.98 & 2.47 & 13.94x & 1.26 & 60.55 & 41.50 & 14.44x & 2.60\\
ICAE & 62.08 & 37.70 & 23.16x & 3.76 & 38.53 & 3.30 & 14.58x & 1.27 & 64.30 & 46.78 & 21.48x & 3.01\\
500Compressor & 64.30 & 39.60 & 23.16x & 3.69 & 40.28 & 3.38 & 14.58x & 1.23 & 69.63 & 50.62 & 21.48x & 3.12\\
QGC & 68.18 & 45.12 & 14.58x & 1.95 & 44.20 & 5.20 & 15.76x & 0.81 & 68.40 & 50.43 & 16.13x & 1.76\\

\midrule
ATACompressor & \textbf{78.44} & \textbf{62.65} &\textbf{24.15x}& 3.86 & \textbf{50.06} & \textbf{8.00}&\textbf{27.36x} & 1.29 & \textbf{71.67} & \textbf{53.00} & \textbf{27.18x}  & 3.14\\
\bottomrule
\end{tabular}
\label{tab:comparison}
\end{table*}

\section{Experiments}
\subsection{Settings}
\subsubsection{Datasets}
\label{4.1.1}
The experiments are based on the three datasets:

\begin{itemize}[itemsep=0.2em, topsep=0.2em]
    \item \textbf{HotpotQA} \cite{yang2018hotpotqa}: HotpotQA is a multi-hop question answering dataset where the answer requires information from more than one document. We use it to evaluate models at the \textbf{document level}, where the LLM needs to aggregate information from multiple docs to generate a correct answer.

    \item \textbf{MSMARCO} \cite{nguyen2016ms}: MSMARCO (Question Answering v2.1) is a high-quality question answering dataset curated by Microsoft. In this study, we employ the dataset to assess models at the \textbf{passage level}, where the LLM is tasked with synthesizing information from relevant passages to produce the correct answer.

    \item \textbf{SQUAD} \cite{rajpurkar2018know}: SQUAD is a question-answering dataset where each question is paired with a passage, and the answer is typically a span of text found within that passage. We utilize SQUAD is structured to assess models at the \textbf{sentence level}, demanding the LLM to aggregate information from sentences to generate a correct answer.
    
\end{itemize}
\vspace{-0.1mm}
\subsubsection{Baselines}
\label{4.1.2}
We use three types of baselines.

\textit{1. No Compression}
\begin{itemize}[itemsep=0.2em, topsep=0.2em]
    \item \textbf{Closed-Book} The LLM directly answers questions without access to any external context. 
    \item \textbf{Original-Context } The LLM answers questions with access to the full external context, using the original uncompressed context without any modifications.
\end{itemize}
\textit{\quad2. Hard Prompt Compression}
\begin{itemize}[itemsep=0.2em, topsep=0.2em]
    \item \textbf{Selective-Context} \cite{li2023compressing}: It leverages self-information computed by an external language model to remove redundant words.
   \item \textbf{LongLLMLingua} \cite{jiang2023longllmlingua}: It uses a language model to assess document importance via question-aware perplexity and applies a coarse-to-fine strategy to remove irrelevant tokens.
\end{itemize}
\textit{\quad3. Soft Prompt Compression}
\begin{itemize}[itemsep=0.2em, topsep=0.2em]
   \item \textbf{AutoCompressor} \cite{chevalier2023adapting}: It fine-tunes an LLM to iteratively compress long contexts into summary vectors. In our experiments, we use the released AutoCompressor-Llama-2-7B-6K model\footnote{https://huggingface.co/princeton-nlp/AutoCompressor-Llama-2-7b-6k} for experiments.
    \item \textbf{ICAE} \cite{ge2023context}: It adopts an autoencoder architecture to compress long contexts into compact memory slots. 
    \item \textbf{500Compressor} \cite{li2024500xcompressor}: Similar to ICAE, the key difference is that it uses the KV representations of the compressed tokens instead of the embeddings. 
    \item \textbf{QGC} \cite{cao2024retaining}: It compresses query-guided document representations into n-grams based on word importance to the query.
\end{itemize}

\subsubsection{Main Evaluation Metrics}
We evaluate downstream QA tasks using F1 score and Exact Match (EM). F1 score balances precision and recall, while EM checks if the predicted answer matches the ground truth exactly. We also compute the compression ratio (CR), the ratio of original to compressed context length, and report inference throughput (TP) on two A100-40G GPUs, including compression and answer generation. In addition, Rouge-L-F, which calculates the harmonic mean of precision and recall based on the longest common subsequence (LCS), is used in \S\ref{5} to evaluate the performance of the pretraining (regeneration) task.

\subsubsection{Implementation Details}
We utilized a 280k dataset (comprising 180k MSMARCO and 100k HotpotQA) from the training sets for pretraining and finetuning. We randomly sampled 5k examples from the test sets of MSMARCO, HotpotQA, and SQUAD for testing. All subsequent experimental results are averages of 5 random samplings, unless otherwise specified. For ATACompressor, following \S \ref{3.3}, the dataset was partitioned into \texttt{<PA></PA>} chunks, with the granularity described in \S \ref{4.1.1}. Specifically, for HotpotQA, each document was treated as a chunk; for MSMARCO, each passage was treated as a chunk; and for SQUAD, each sentence was treated as a chunk. The chunks were concatenated to form a preprocessed context, which was then combined with the query to create the input text. Other models used the same chunked data as ATACompressor. To facilitate comparison with baselines, the maximum input length was set to 600 tokens (as many baselines conduct their main experiments using input lengths around 500 tokens and resource limitation), and only inputs below this limit were retained during dataset construction.
In our experiments, we used LLaMA-2-7B and Qwen-2-7B as backbone models, training all models with open-source code unless specified in \S \ref{4.1.2}. For ATACompressor, we set the hyperparameter $\lambda$ in Eq. (\ref{eq:pretrain}) to $10^{-4}$ and $\delta$ in Eq. (\ref{eq:huber}) to 10 in pretraining. The policy ratio $r$ in Eq. (\ref{eq:policy}) was randomly chosen from \{1, 5, 10, 20, 50\} for each batch during pretraining, and fixed at 10 for finetuning and reporting (unless otherwise specified). The maximum number of compressed tokens $k_{max}$ was set to 8 for both training and inference \footnote{Other details can be found at https://anonymous.4open.science/r/ATACompressor}.
\subsection{Main Results}
Table \ref{tab:comparison} shows the performance of various methods across three benchmark datasets. ATACompressor consistently outperforms all other methods in both task performance (F1 and EM scores) and compression efficiency. It achieves the highest compression ratios while maintaining excellent throughput, highlighting its effectiveness and efficiency in long-context compression.

First of all, ATACompressor demonstrates significant advantages over non-compression methods, effectively alleviating the "lost in the middle" issue by focusing on key information within lengthy contexts. 
Furthermore, ATACompressor achieves the highest relative improvement on the HotpotQA dataset, which has the largest granularity, longer overall text, and longer single-chunk length. The relatively larger amount of irrelevant text in this dataset highlights the advantage of ATACompressor's selective compression, demonstrating its strong ability to handle long-context scenarios with the selective encoder.
Results on MSMARCO and SQUAD further demonstrate ATACompressor’s capability to selectively preserve task-relevant information across varying context lengths, maintaining both high performance and compression efficiency. Another key highlight of ATACompressor is its high compression rate. This is made possible by its adaptive resource allocation feature: by analyzing the length of relevant content, ATACompressor dynamically adjusts the compression rate to optimize resource utilization. 
As a result, it consistently achieves high compression ratios across datasets, effectively compressing long contexts into compact forms while preserving information quality.
This is particularly obvious in datasets with longer contexts, such as HotpotQA, where ATACompressor achieves both the highest compression efficiency and superior task performance. 
Furthermore, its competitive throughput demonstrates strong efficiency, making it well-suited for real-time or large-scale applications. Finally, experiments across different models highlight ATACompressor’s adaptability, as it achieves significant performance improvements regardless of the underlying architecture.

\subsection{Ablation Study}
\begin{table}[!t]
\centering
\caption{The ablation study results on MSMARCO using LLAMA-2-7B. Here, $k$ represents the number of compressed tokens or average.}
\vspace{-5pt}  
\begin{tabular}{lcccc}
\hline
\textbf{Methods} & \textbf{F1} & \textbf{EM} & \textbf{CR(k)} & \textbf{TP} \\
\hline
ATAcompressor & \textbf{50.06} & \textbf{8.00} & \textbf{27.36x (4.18)} & 1.29 \\
\hline
\textit{w/o AAC} & 47.52 & 7.34 & 19.06x (6.00) & 1.29 \\
\textit{w/o Selective} & 40.83 & 3.51 & 24.59x (4.65) & 1.29 \\
\hline
\end{tabular}
\vspace{-2.0em} 
\label{tab:ablation}
\end{table}

As shown in Table \ref{tab:ablation}, we conduct two types of ablation studies to investigate the effectiveness of our approach:

(1) w/o adaptive allocation controller (AAC): This variant replaces the adaptive allocation controller with a fixed number of compressed tokens.
As a result, the method exhibits a decrease in both performance and compression ratio, indicating a less efficient use of resources. This highlights the importance of the adaptive allocation controller: it can preserve essential information and optimize resource utilization across varying context lengths by dynamically adjusting the compression rate based on task. The throughput (TP) is similar across the three cases, as mentioned in \S\ref{3.2}, where the selective encoder and AAC work in parallel.

(2) w/o Selective: This variant removes the selective compression step, compressing all context tokens instead of focusing on task-relevant ones. In this case, the adaptive allocation controller does not estimate the length of relevant content, and instead uses the full context length to adjust the compression rate. As shown in Table \ref{tab:ablation}, the removal of the selective encoding mechanism leads to a noticeable decline in both performance and compression ratio. This indicates that without the selective encoder, the compression fails to prioritize task-specific information, resulting in suboptimal performance and reduced compression efficiency.


\section{Analysis}
\label{5}
In this section, we conduct a series of experiments to further investigate the effectiveness and efficiency of ATACompressor. We choose ICAE and 500Compressor as baselines due to their strong performance in both effectiveness, efficiency, and their broad applicability. We adjust the average number of compressed tokens \( k \) for ATACompressor and ATACompressor-w/o-Selective by directly modifying the policy ratio \( r \), while for other methods, \( k \) is adjusted by training models with different values of \( k \). 
\begin{figure}[!t]
    \centering
    \begin{subfigure}[b]{0.4\textwidth}
        \centering
        \includegraphics[width=\textwidth]{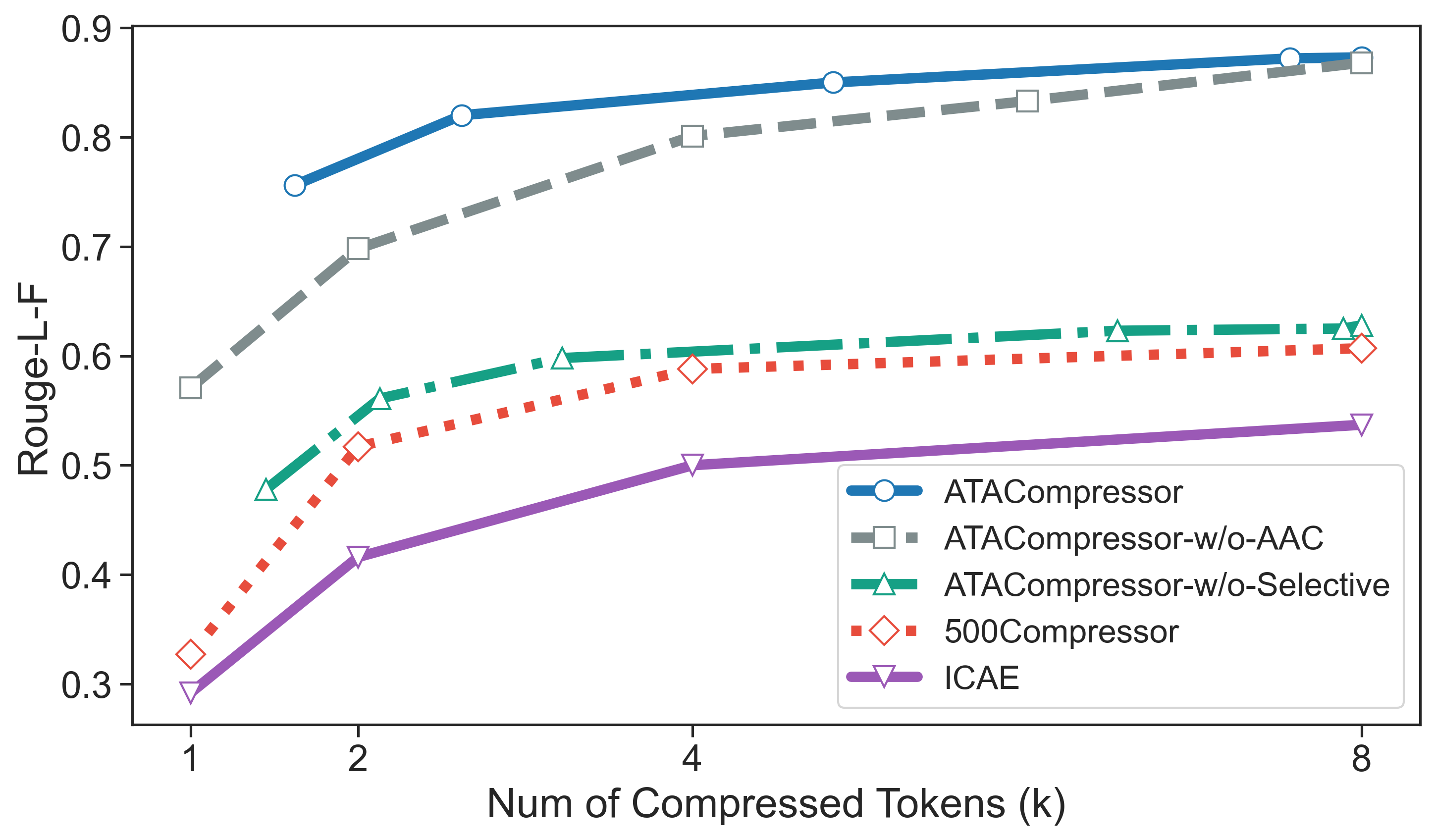} 
        \caption{Pretraining Results (Regeneration)}
        \label{fig:sub1}
    \end{subfigure}
    \hfill
    \begin{subfigure}[b]{0.4\textwidth}
        \centering
        \includegraphics[width=\textwidth]{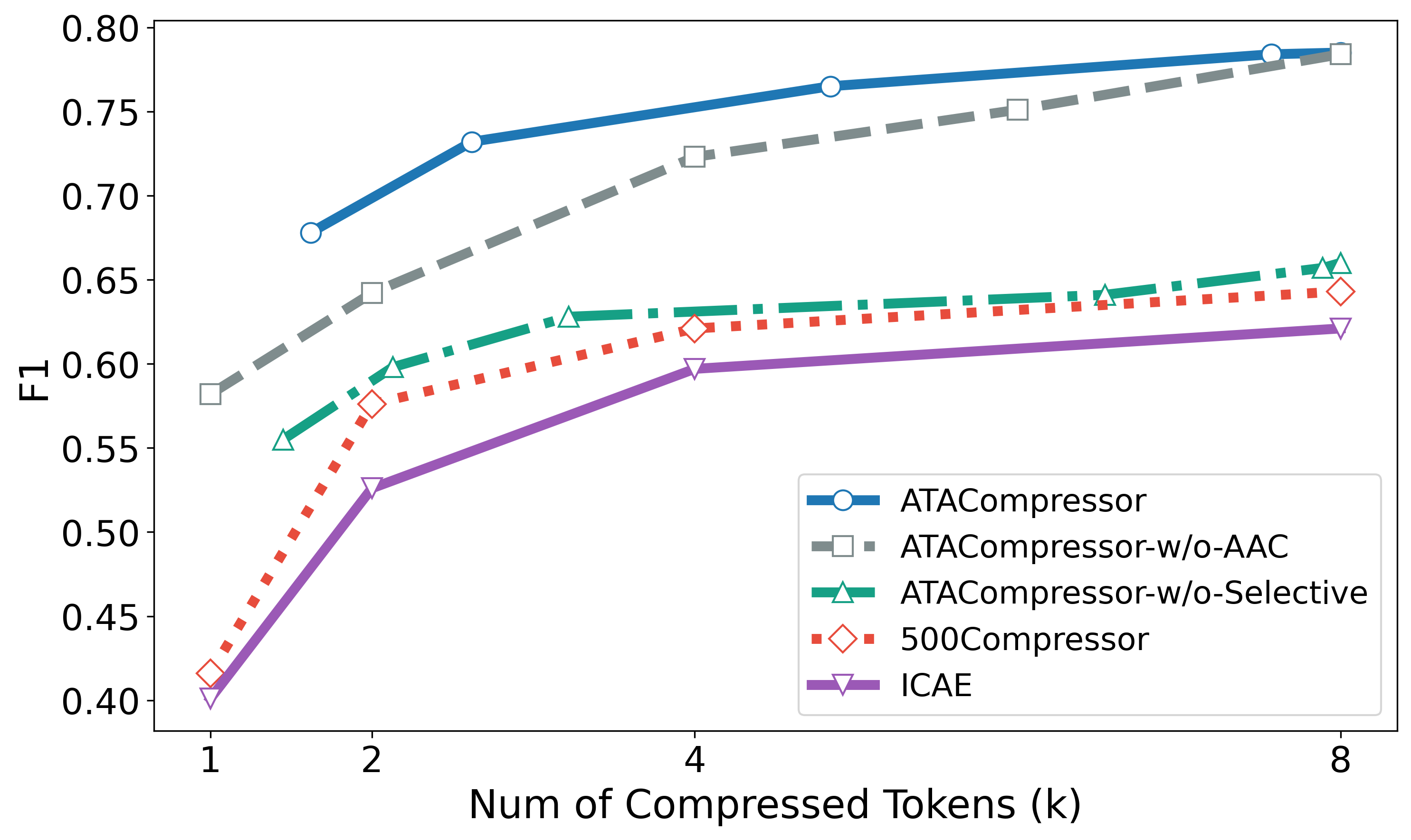} 
        \caption{Finetuning Results (QA)}
        \label{fig:sub2}
    \end{subfigure}
    \caption{Performance on pretraining (regeneration) and finetuning (QA) tasks with varying numbers of compressed tokens using the LLAMA-2-7B model on HotpotQA.}
    \label{fig:number}
    \vspace{-2.0em}
\end{figure}
\subsection{Impact of the Compressed Tokens' Number}
\label{5.1}
The analysis in Figure \ref{fig:number} on the impact of the number of compressed tokens \( k \) demonstrates ATACompressor’s robustness. As \( k \) decreases, ATACompressor shows a smaller performance drop compared to other methods, highlighting its ability to handle varying compression levels with minimal performance loss.
While all methods experience some degradation as \( k \) reduces, ATACompressor maintains relatively high task performance, even under tighter compression. In contrast, ICAE and 500Compressor suffer sharp performance declines, emphasizing ATACompressor’s ability to efficiently preserve critical content with fewer tokens.
Additionally, both ATACompressor-w/o-AAC and ATACompressor-w/o-Selective also see performance drops as \( k \) decreases, underscoring the combined importance of the selective encoder and adaptive allocation controller. Together, these components enable ATACompressor to maintain competitive performance across varying \( k \), demonstrating its ability to optimize efficiency and accuracy under different compression limitations.
Additionally, for both ATACompressor and ATACompressor-w/o-Selective, the number of compressed tokens $k$ is controlled by adjusting the policy ratio $r$. A larger $r$ leads to a smaller average $k$, which typically results in higher compression but worse task performance. This highlights that, in practical deployment, the value of $r$ should be chosen based on both available computational resources and task performance requirements, allowing the system to flexibly balance efficiency and effectiveness.




\begin{figure}[!t]
    \centering
    \includegraphics[width=0.4\textwidth]{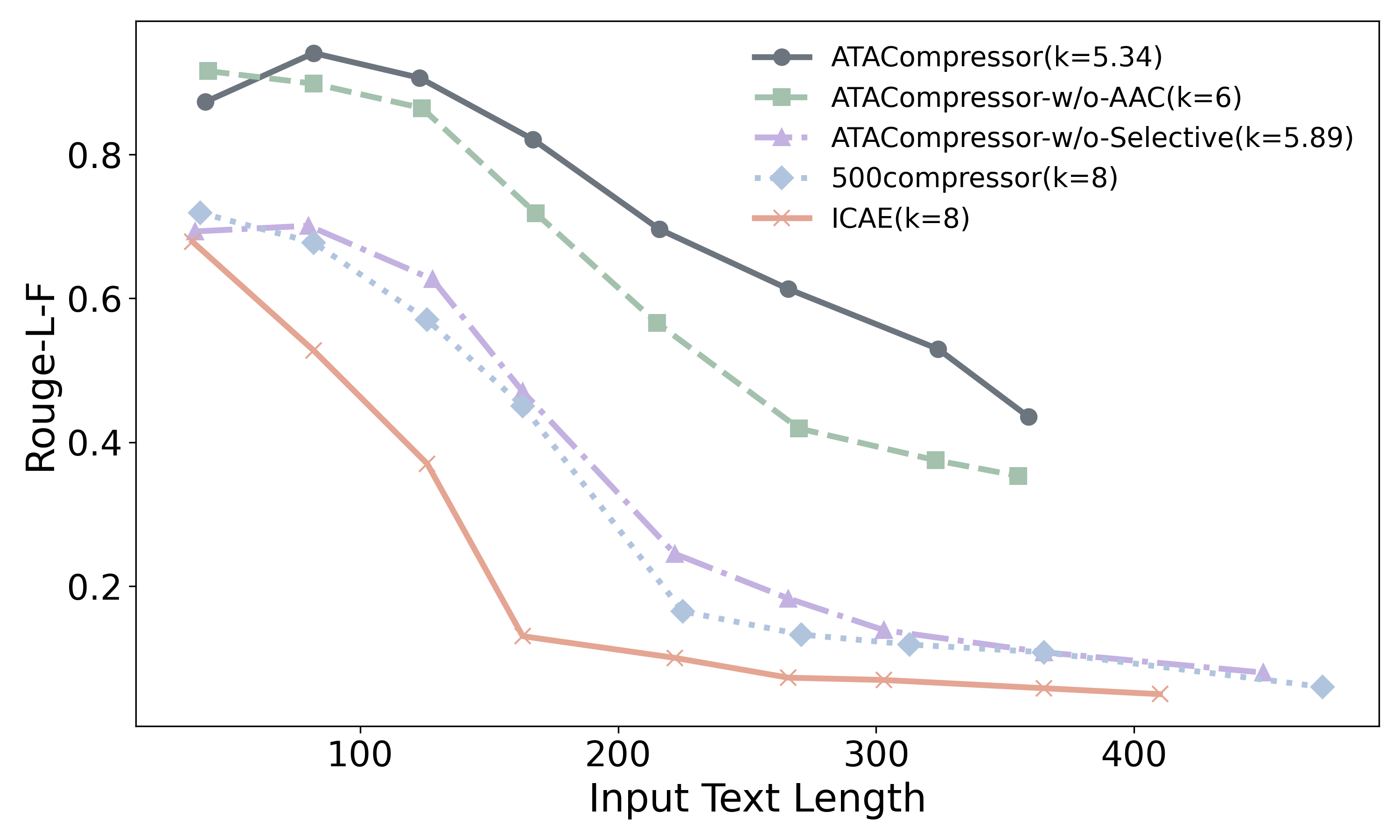} 
   \caption{Performance on pretraining (regeneration) across varying input text lengths using the Qwen-2-7B. $k$ represents the number of compressed tokens or average.}
    \label{fig:length} 
    \vspace{-2.0em}
\end{figure}
\subsection{Performance Across Input Text Lengths}
Figure \ref{fig:length} demonstrates the effectiveness of ATACompressor in handling varying input text lengths. The relatively poor performance of ATACompressor on shorter texts is due to our policy, which results in a very small number of compressed tokens being used when the text length is low. In practical applications, a more refined strategy can be adopted to address this issue, such as setting a reasonable minimum compressed token number. However, overall, ATACompressor exhibits strong robustness for longer texts, significantly outperforming 500Compressor and ICAE as the input text length increases. This adaptability enables ATACompressor to more effectively preserve key information across different context lengths.
ATACompressor-w/o-AAC, ICAE, and 500Comprssor compresses texts of varying lengths into a fixed number of compressed tokens, which results in good performance for shorter texts. However, as the text length increases, performance rapidly declines. In comparison, ATACompressor-w/o-Selective shows a similar trend to ATACompressor but lacks the selective mechanism, leading to significantly lower performance. This further underscores the combined importance of the selective encoder and adaptive allocation controller in maintaining ATACompressor’s superior performance.



\begin{table*}[!t]
\centering
\small
\caption{Case study of ATACompressor and ICAE on Qwen-2-7B. \colorbox{red!50}{red} highlights errors in key information, typically referring to incorrect statements directly impacting the answer. \colorbox{yellow!30}{yellow} indicates mistakes in less critical details that do not directly affect the core answer. \colorbox[rgb]{0.498, 0.741, 0.557}{green} denotes hallucinations, where the text contains information not present in the original source. \colorbox[rgb]{0.6, 0.8, 1}{blue} signals information loss, meaning less important content from the original text is missing. Finally, \colorbox{brown!50}{brown} indicates paraphrasing, where the original content is reworded without changing its meaning. Also, $k$ represents the number of compressed tokens or average.
}
\begin{tabular}{|c|}
\hline
\textbf{Original Text (The input to ICAE does not include the labels \texttt{<PA>} and \texttt{</PA>}.)} \\
\hline
\begin{minipage}[t]{1.0\textwidth}
\raggedright
<PA> The Normans (Norman: Nourmands; French: Normands; Latin: Normanni) were the people who in the 10th and 11th centuries gave their name to Normandy, a region in France. </PA> 
<PA> They were descended from Norse ("Norman" comes from "Norseman") raiders and pirates from Denmark, Iceland and Norway who, under their leader Rollo, agreed to swear fealty to King Charles III of West Francia. </PA>  
<PA> Through generations of assimilation and mixing with the native Frankish and Roman-Gaulish populations, their descendants would gradually merge with the Carolingian-based cultures of West Francia. </PA> 
<PA> The distinct cultural and ethnic identity of the Normans emerged initially in the first half of the 10th century, and it continued to evolve over the succeeding centuries. </PA> 
\end{minipage} \\
\hline
\end{tabular}

\vspace{0.1cm} 

\begin{tabular}{|p{3cm}|p{6cm}|p{8.05cm}|}
\hline
\textbf{Question}& \textbf{ATACompressor Regeneration Output (The value of \( k \) is 2, 4, and 2 for the following three questions, respectively.)} &\textbf{ICAE Regeneration Output (k=8)} \\
\hline
\textbf{Q1: In what country is Normandy located?}  &The Normans (\colorbox[rgb]{0.6, 0.8, 1}{Norman}; French: Normands; Latin: Normanni) were the people who in the 10th and 11th centuries gave their name to Normandy\colorbox{yellow!40}{;} a region in France. & \multirow{3}{8.05cm}{The Normans (\colorbox{yellow!40}{Normand;} Nourmands\colorbox{yellow!40}{:} \colorbox[rgb]{0.6, 0.8, 1}{French} \colorbox{yellow!40}{;} Latin: Normanni) were the people who \colorbox[rgb]{0.498, 0.741, 0.557}{were the Normans} in the 10th and 11th centuries \colorbox{brown!50}{who} gave their name to Normandy, a region in France. \colorbox[rgb]{0.6, 0.8, 1}{They} descended from Norse ("Norman" comes from "Norseman") raiders and pirates from Denmark, Iceland and Norway who, under their leader Rollo, agreed to swear fealty to King Charles III of West Francia. Through generations of assimilation and mixing \colorbox{yellow!40}{of} the \colorbox{yellow!40}{Norman and Frankish cultures}, their \colorbox[rgb]{0.498, 0.741, 0.557}{first- and second-generation} descendants would \colorbox{yellow!40}{inherit the Roman-Gallic cultural identity of the} \colorbox{yellow!40}{Carolingian Normans}. The \colorbox{brown!50}{distinctive} cultural and ethnic \colorbox[rgb]{0.6, 0.8, 1}{identity} emerged \colorbox{yellow!40}{gradually over} the first half of the \colorbox{red!50}{11th} century, 
\colorbox[rgb]{0.498, 0.741, 0.557}{and it was the native Frankish culture that initially dominated} \colorbox[rgb]{0.498, 0.741, 0.557}{ Western Europe.}} \\
\cline{1-2}
\textbf{Q2: From which countries did the Norse originate?}  &The Normans (\colorbox[rgb]{0.6, 0.8, 1}{Norman}; French: Normands; Latin: Normanni) were the people who in the 10th and 11th gave their name to Normandy\colorbox{yellow!30}{;} a region in France. They were descended from Norse ("Norman" comes from "Norseman") raiders and pirates from Denmark, Iceland, and Norway who, \colorbox{brown!50}{under the leadership of Rollo}, agreed to swear fealty to King Charles III of West Francia. & \\
\cline{1-2}
\textbf{Q3: What century did the Normans first gain their separate identity?} & The \colorbox{brown!50}{distinctive} cultural and ethnic identity of the Normans emerged initially in the first half of the 10th century, and it continued to evolve over the succeeding centuries. \quad\quad\quad\quad\quad\quad\quad\quad\quad\quad\quad\quad\quad\quad\quad\quad \quad\quad\quad\quad\quad\quad& \\
\hline
\end{tabular}
\label{tab:case_study}
\end{table*}

\subsection{Performance of the Probe}

\begin{figure}[t!]
    \centering
    \includegraphics[width=0.35\textwidth]{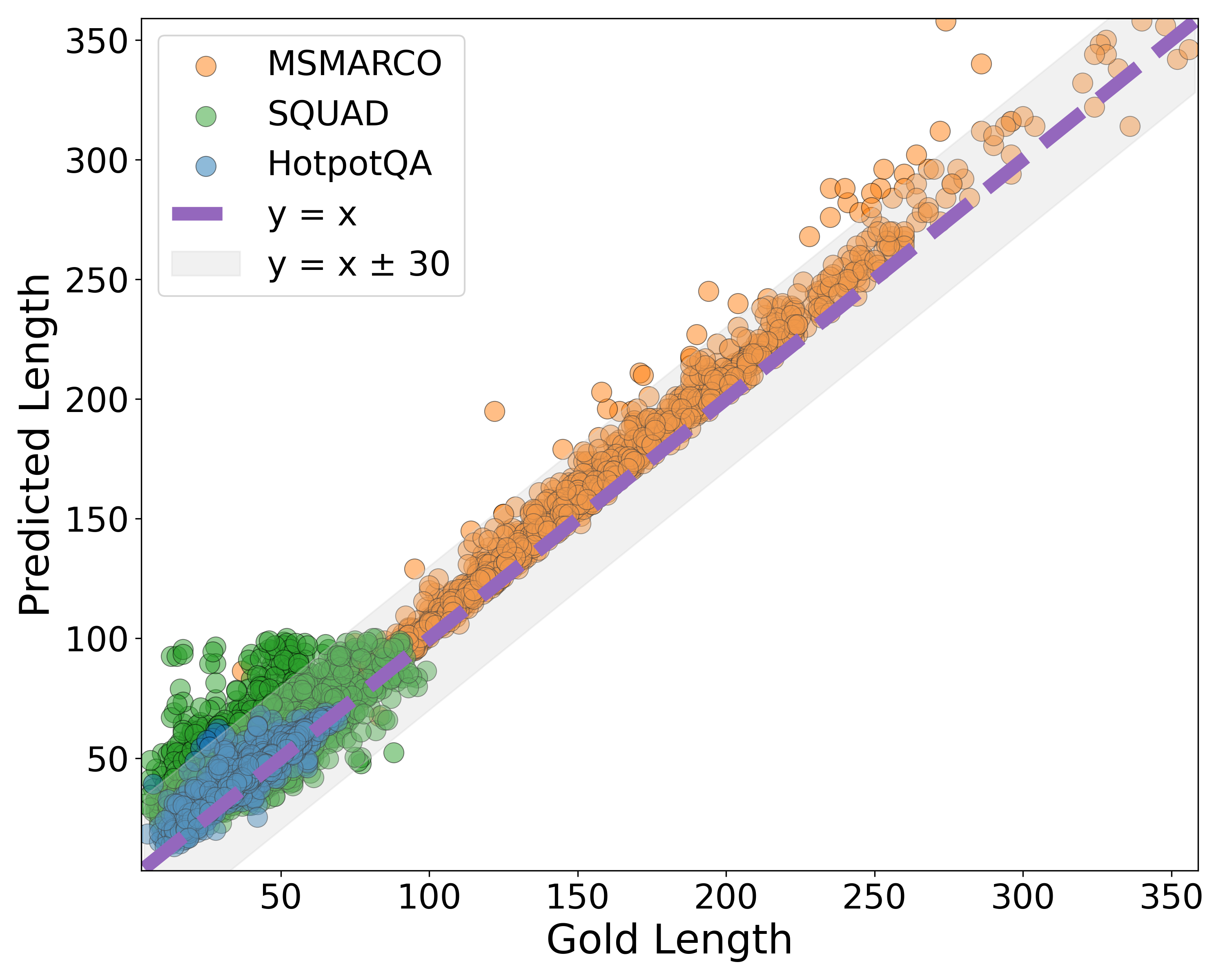} 
    \caption{Comparison of gold (\(L_{Rel}\)) and predicted lengths (\(\hat{L}_{Rel}\)) across three datasets on Qwen-2-7B.}
    \label{fig:probe} 
    \vspace{-2.0em}
\end{figure}
Figure \ref{fig:probe} shows that the adaptive allocation controller demonstrates good prediction accuracy across all three datasets. Furthermore, we argue that determining whether a sentence is necessary for answering a query is inherently more challenging than deciding whether an entire document is relevant. Sentence-level judgments are more uncertain because they lack sufficient contextual information, while document-level judgments benefit from a broader context, allowing for a more accurate identification of relevant information. This explains why HotpotQA, despite having the longest average text length, achieves relatively low prediction error. Its document-level granularity enables the encoder to more easily identify which chunks are relevant, leading to more accurate text length estimations. In contrast, SQUAD's sentence-level granularity introduces higher uncertainty, resulting in a higher prediction error. These findings suggest that the selective encoder operating on higher granularity levels can leverage a stronger global understanding of the context, which helps reduce uncertainty and improves prediction accuracy.

\subsection{Computational Efficiency}
\label{5.4}
\begin{table}[h!]
    \centering
    \small 
    \caption{QA task efficiency evaluated on the HotpotQA using the LLaMA2-7B model on two A100-40G GPUs. $k$ represents the number of compressed tokens or average.}
    \label{tab:decoding_efficiency}
    \begin{tabular}{p{2.3cm}ccc}
        \toprule
        \textbf{Method} & $k$ & \textbf{Inference Time (ms)} & \textbf{GPU Mem. (GB)} \\
        \midrule
        Closed Book & -   & \textbf{157.00} & \textbf{18.79}  \\
        Original-Context & -  & 826.46  & 21.58 \\
        \midrule
        Selective-Context & -   & 819.69 & 23.82 \\
        LongLLMLingua & -  & 757.56  & 33.56 \\
        \midrule
        Autocompressor & 15.33   & 337.83 & 25.56 \\
        ICAE & 8.00   & 265.10  & 23.97 \\
        500Compressor & 8.00   & 270.29  & 24.32  \\
        QGC   & 12.80   & 510.18  & 35.44   \\
        \midrule
        ATACompressor-w/o-AAC  & 8.00   & 254.18  & 24.30  \\
        ATACompressor-w/o-Selective   & 7.91   & 255.10  & 28.42  \\
        \midrule
        ATACompressor   &  7.59  & 255.08 & 28.66   \\
                       & 1.62   & 254.89  & 28.49 \\

        \bottomrule
    \end{tabular}
    \vspace{-2em}
\end{table}
Table \ref{tab:decoding_efficiency} compares the efficiency of different methods for the QA task in terms of inference time and GPU memory cost. 
It shows that ATACompressor demonstrates excellent efficiency, with low inference time and GPU memory usage. The performance remains stable across a small range of compressed tokens \(k\). Compared to Orginal-Context method, ATACompressor significantly reduces inference time. 
When compared to QGC, which uses a soft prompt framework for query-based compression, ATACompressor achieves lower inference times and GPU memory usage, demonstrating its efficiency.

\subsection{Case Study}
Table \ref{tab:case_study} presents the results of a case study comparing ICAE and ATACompressor. Unlike ICAE, which performs full-text compression, ATACompressor selectively compresses relevant context according to task-specific needs, ensuring critical information is preserved and reducing the risk of key errors. For instance, in Question 3, ICAE Introduced a critical error by incorrectly stating "over the first half of the 11th century" instead of the correct text "in the first half of the 10th century". Additionally, ATACompressor employs adaptive compression, dynamically adjusting token usage based on the length of relevant content. This mechanism optimizes resource efficiency while maintaining high performance across tasks.
\subsection{Cross‑Task Generalization on LongBench}\label{sec:cross‑task}
To study the cross‑task generalization of ATACompressor, we evaluate it on two representative tasks from \textsc{LongBench} \cite{bai2023longbench}. We chunk the Summary task at the sentence level and the Few‑shot task at the demonstration‑pair level, using only inputs shorter than 2048 tokens.
Results in Table \ref{tab:longbench_summary_fewshot_pred} demonstrate that ATACompressor not only attains state‑of‑the‑art effectiveness but also maintains comparatively high efficiency. This demonstrates that ATACompressor performs well across a wide variety of unseen tasks, highlighting its strong potential for cross-task generalization.

\begin{table}[ht]
\centering
\small
\caption{Performance on \textit{Summary} and \textit{Few‑shot} tasks of LongBench using Qwen-2-7B model.}
\begin{tabular}{@{}lcccccc@{}}
\toprule
\textbf{Methods} & \multicolumn{3}{c}{\textbf{Summary}} & \multicolumn{3}{c}{\textbf{Few‑shot}} \\
\cmidrule(lr){2-4} \cmidrule(lr){5-7}
& Score & CR & TP & Score & CR & TP \\
\midrule\midrule
Closed‑book            & 10.43 & --           & \textbf{3.04} & 41.02 & --           & \textbf{3.12} \\
Original‑Context       & 26.32 & 1.00$\times$ & 0.86          & 58.71 & 1.00$\times$ & 0.92 \\
Selective‑Context      & 22.13 & 3.82$\times$ & 0.88          & 48.35 & 2.93$\times$ & 0.90 \\
LongLLMLingua          & 29.21 & 5.24$\times$ & 0.93          & 60.14 & 4.13$\times$ & 0.97 \\
ICAE                   & 30.09 & 21.38$\times$& 1.88          & 62.38 & 15.64$\times$& 1.95 \\
500Compressor          & 31.43 & 21.45$\times$& 1.81          & 63.77 & 15.59$\times$& 1.87 \\
QGC                    & 33.23 & 15.18$\times$& 1.13          & 66.18 & 11.54$\times$& 1.19 \\
\textbf{ATACompressor} & \textbf{37.68} & \textbf{24.79$\times$} & 1.94 & \textbf{71.83} & \textbf{18.37$\times$} & 1.98 \\
\bottomrule
\end{tabular}
\label{tab:longbench_summary_fewshot_pred}
\vspace{-2em}
\end{table}

\section{Conclusion}
We present ATACompressor, a scalable solution for efficient long-context processing in LLMs. By using a selective encoder and adaptive allocation controller, it dynamically compresses context based on task needs for optimal efficiency. Experiments show ATACompressor surpasses existing methods, achieving higher compression and throughput while retaining essential information, demonstrating its strength in improving LLM performance.

\bibliographystyle{ACM-Reference-Format}
\balance
\bibliography{sample-base}
\end{document}